\def\year{2020}\relax
\documentclass[letterpaper]{article}
\usepackage{aaai20}
\usepackage{times}
\usepackage{helvet}
\usepackage{courier}
\usepackage{url}
\usepackage{graphicx}
\usepackage{afterpage,lipsum}
\usepackage{amsmath}
\usepackage{array}
\usepackage{hyperref}
\usepackage{xcolor}
\newcommand{\link}[1]{{\color{blue}\href{#1}{#1}}}
\usepackage{pdfpages}
\usepackage{booktabs}
\newcommand{\citep}{\cite}
\title{Fractional Skipping: Towards Finer-Grained Dynamic CNN Inference}
\author{
Jianghao Shen\textsuperscript{\rm 1,\rm 2},
Yonggan Fu\textsuperscript{\rm 1},
Yue Wang\textsuperscript{\rm 1},
Pengfei Xu\textsuperscript{\rm 1},
Zhangyang Wang\textsuperscript{\rm 2},
Yingyan Lin\textsuperscript{\rm 1}\\
\textsuperscript{\rm 1}Rice University
\textsuperscript{\rm 2}Texas A \& M University \\
nie@tamu.edu, yf22@rice.edu, yw68@rice.edu, px5@rice.edu, atlaswang@tamu.edu, yingyan.lin@rice.edu}
\begin{document}
\maketitle

\begin{abstract}
While  increasingly  deep  networks  are  still  in  general  desired  for  achieving  state-of-the-art  performance,  for  many specific inputs a simpler network might already suffice. Existing works exploited this observation by learning to skip convolutional layers in an input-dependent manner. However, we argue their binary decision scheme, i.e., either fully executing or completely bypassing one layer for a specific input, can be enhanced by introducing finer-grained, “softer” decisions. We therefore propose a Dynamic Fractional Skipping (DFS) framework. The core idea of DFS is to hypothesize layer-wise quantization  (to different  bitwidths)  as  intermediate  ``soft'' choices  to  be  made  between  fully  utilizing  and  skipping  a layer. For each input, DFS dynamically assigns a bitwidth to both weights and activations of each layer, where fully executing and skipping could be viewed as two ``extremes'' (i.e., full bitwidth and zero bitwidth). In this way, DFS can “fractionally” exploit a layer’s expressive power during input-adaptive inference, enabling finer-grained accuracy-computational cost trade-offs. It presents a unified view to link input-adaptive layer skipping and input-adaptive hybrid quantization. Extensive experimental results demonstrate the superior tradeoff between computational cost and model expressive power (accuracy) achieved by DFS. More visualizations also indicate a smooth and consistent transition in the DFS behaviors, especially the learned choices between layer skipping and different quantizations when the total computational budgets vary, validating our hypothesis that layer quantization could be viewed as intermediate variants of layer skipping. Our source code and supplementary material are available at \link{https://github.com/Torment123/DFS}.
\end{abstract}

\section{Introduction}
Although convolutional neural networks (CNNs) have show state of the art performance in many visual perception tasks \citep{krizhevsky2012imagenet,taigman2014deepface}, the high computational cost has limited their application in resource constrained platforms such as drones, self-driving cars, wearables and many more. The growing demand of unleashing the intelligent power of CNN into these devices has posed unique challenges in developing algorithms that enables more computationally efficient inference of CNNs. Earlier resource-efficient implementations assumed that CNNs are first compressed before being deployed, thus being ``static" and unable to adjust their own complexity at inference. Later on, a series of works \citep{figurnov2017spatially,wang2018skipnet} pointed out that the continuous improvements in accuracy, while significant, are marginal compared to the growth in model complexity. This implies that computationally intensive models may only be necessary to classify a handful of difficult inputs correctly, and they might become ``wasteful'' for many simple inputs. 

Motivated by this observation, several works have tackled the problem of input-dependent adaptive inference, by dynamically bypassing unnecessary computations on the \textit{layer level}, i.e., selectively executing a subset of layers 
\citep{figurnov2017spatially,wu2018blockdrop}. 
However, the \textit{binary} decision scheme of either executing a layer fully, or skipping it completely, leaves no room for intermediate options. \textbf{We conjecture that} finer-grained dynamic execution options can contribute to better calibrating the inference accuracy of CNNs w.r.t. the complexity consumed. 

On a separate note, CNN quantization appears to exploit model redundancy at the finest level - by reducing the bitwidth of the \textit{element-level} numerical representations of weights and activations. Earlier works \citep{han2015deep,zhu2016trained} presented to quantize all layer-wise weights and activations to the same low bitwidth, yet ignored the fact that different layers can have different importance. The latest work \citep{wang2019haq} learned to assign different bitwidths for each layer. However, no work has yet discussed an input-adaptive, layer-wise bitwidth allocation at the inference time, not to mention linking between quantization with dynamic inference. 

In an effort to enable finer-grained dynamic inference beyond ``binary'' layer skipping, we propose a Dynamic Fractional Skipping (DFS) framework, that treats layer quantization (to different bitwidths) as softer, intermediate versions of layer-wise skipping. Below are our contributions:

\begin{itemize}
\item We propose to link two efficient CNN inference mindsets: \textit{dynamic layer skipping} and \textit{static quantization}, and show that they can be unified into one framework. Specifically, DFS considers a quantized layer to be a ``fractionally executed'' layer, in contrast to either a fully executed (selected) or non-executed (bypassed) layer in the existing layer skipping regime. In this way, DFS can more flexibly calibrate the trade-off between the inference accuracy and the total computational costs. 

\item We introduce \textit{input-adaptive quantization} at inference \textbf{for the first time}. Based on each input, DFS learns to dynamically assign different bitwidths to both weights and activations of different layers, using a two-step training procedure. That is in contrast to \citep{wang2019haq} that learns layer-wise bit allocation during training, which is then fixed for inference regardless of inputs. The existing layer skipping could be viewed as DFS's coarse-grained version, i.e., allowing only to select between full bits (executing without quantization) and zero bit (bypassing). 

\item We conduct extensive experiments to illustrate that DFS strikes a better computational cost and inference accuracy balance, compared to dynamic layer skipping and other relevant competitors. Moreover, we visualize the skipping behaviors of DFS when varying the total inference computations in a controlled way, and observe a smooth transition from selecting, to quantizing, and to bypassing layers. The observation empirically supports our conjecture that layer quantization can be viewed as soft and intermediate variants of layer skipping.

\end{itemize}

\section{Related Works}

\textbf{Model Compression.} Model compression has been widely studied to speedup CNN inference by reducing model size \citep{wu2018deep}. Existing works focus on pruning unimportant model weights, or quantizing the model into low bitwidths.

\underline{Pruning}:
There has been extensive studies on model pruning in different granularities. \citep{han2015deep,han2015learning} reduce the redundant parameters by performing element-wise weight pruning. Coarser-grained channel level pruning has been explored in \citep{yu2017scalpel,liu2017learning,he2017channel} by enforcing group sparsity. \citep{wen2016learning} exploits parameter redundancy in a multi-grained manner by grouping weights into structured groups during pruning, each with a Lasso regularization. \citep{xu2018hybrid} proposes a hybrid pruning by performing element-wise pruning on top of the filter-wise pruned model. \citep{kim2018nestednet} performs multi-grained model pruning by adding explicit objectives for different levels. \citep{cheng2017survey} presents a comprehensive review on pruning techniques. These methods are applied to well-trained networks and do not dynamically adjust the model complexity conditioned on the input. 

\underline{Network Quantization}:
Quantizing network weights and activations has been proven to be an effective approach to reduce the memory and computational budgets. Most of the existing works quantize the model to varied bitwidths with a marginal accuracy loss. \citep{rastegari2016xnor} binarized each convolution filter into \big\{-\textit{w}, +\textit{w}\big\}. \citep{zhou2016dorefa} used one bit for network weights and two bits for activations. \citep{jacob2018quantization} made use of 8-bit integers for both weights and activations. With the recent development of hardware design, it becomes possible to use flexible bitwidths for different layers \citep{wang2019haq}. \citep{han2015deep} determines the layer-wise bit allocation policy based on domain experts; \citep{wang2019haq} further enhanced the idea by automating the decision process with a reinforcement learning method. These works either empirically find fixed bitwidths or automatically learn fixed layer-wise bit allocation regardless of input, ignoring that the importance of each layer may vary with different inputs. Our proposed DFS models are orthogonal to existing static quantization methods.

\textbf{Dynamic Inference.} While model compression presents ``static'' solutions, i.e., the compressed models cannot adaptively adjust their complexity at inference, for improving inference efficiency, the recently developed  dynamic  inference methods  offer a different option to execute partial inference, conditioned on input complicacy or resource constraints.

\underline{Dynamic Layer Skipping}: Many works \citep{wu2018blockdrop,wang2018skipnet,NIPS-18} have formulate dynamic inference as a sequential decision problem, and selectively bypass subsets of layers in the network conditioned on the input. The common approach of these works is to use gating networks to determine the layer-wise skipping policy in ResNet-style models \citep{he2016deep}, which is inherently suitable for skipping design due to its resiudal structure. The work of \textit{SkipNet} uses a hybrid learning algorithm that sequentially performs supervised pretraining and reinforcement fine-tuning, achieving better resource saving and accuracy tradeoff than existing static model compression methods \citep{wang2018skipnet}. BlockDrop \citep{wu2018blockdrop} uses a decision network to train a global skipping policy for residual blocks, and \citep{liu2018dynamic} trains separate control units for the execution policy of sub parts of the network. In these works, a layer will either be executed fully or skipped completely, leaving no space for any intermediate options. We show that by adding ``softer" intermediate quantization options between the two extremes, the DFS framework exploits the layer's expressive power in a finer granularity, achieving a better accuracy than layer skipping methods under the same computational cost.

\underline{Dynamic Channel Selection/Pruning}:
Since layer skipping only works well in network architectures with residual connections, channel pruning methods have been developed to exploit the redundancy in CNNs at a finer level. \citep{lin2017runtime} formulates the channel pruning problem as a Markov decision process, and apply RNN gating network to determine which channel to prune conditioned on the input. GaterNet \citep{chen2018gaternet} uses a separate network to calculate the channel activation strategy. The slimmable neural network \citep{yu2018slimmable} trains the network with varied layer widths, and adjust channel number during inference to meet resource budgets. \cite{teja2018hydranets} selectively executes branches of network based on input. Compare to quantization, the idea of channel selection exploits fine-grained model redundancy in the channel level, which is orthogonal to our method, and can potentially be combined with our framework to yield further resource savings. 

\underline{Early Exiting}: In many real world applications, there are strict resource constraints: the networks should hence allow for ``anytime prediction'' and be able to halt the inference whenever a specified budget is met. A few prior works enable CNNs with ``early exit" functions. \citep{teerapittayanon2016branchynet} adds additional branch classifiers to the backbone CNNs, forcing a large portion of inputs to exit at the branches in order to meet resource demands. \citep{huang2017multi} further boosts the performance of early exiting by aggregating features from different scale for early prediction. The early exiting works have been developed for resource-dependent inference, which is orthogonal to our input-dependent inference, and the two can be combined to yield further resource savings.

\section{The Proposed Framework}\label{method}
In resource constrained platforms, the ideal efficient CNN inference should save as much resource as possible without non-negligible accuracy degradation. This requires the algorithm to maximally take advantage of the model's expressive power, while dropping any redundant parts. Existing works like SkipNet exploit the model redundancy on the layer level, the binary decision of either executing a layer fully or skipping it completely makes it impossible to make use of the layer's representational power in any finer levels. In contrast, CNN quantization exploits the model redundancy in the finest level - by reducing the bitwidth of the numerical representation of weights and activations. Thus, a natural thought is to use bitwidth options to fill in the gap between the binary options of layer skipping, striking an optimal tradeoff between computational cost and accuracy.

We hereby propose a Dynamic Fractional Skipping (\textbf{DFS}) framework that combines the following two schemes into one continuous fine-grained decision spectrum:

\begin{itemize}
    \item \textbf{Input-Dependent Layer Skipping.} On the coarse-grained level, the ``executed" option of the layer skip decision is equivalent to the \textbf{full bitwidth} option of layer quantization in the DFS framework, and the ``skip" option is equivalent to a \textbf{zero-bit} option of layer quantization.
    \item \textbf{Input-Dependent Network Quantization.} On the fine-grained level, any lower than full bitwidth execution can be viewed as ``fractionally" executing a layer, enabling the model to take advantage of the expressive power of the layer in its low bitwidth version.

\end{itemize}

To our best knowledge, DFS is the first attempt to unify binary layer skipping design and one alternative of its intermediate ``soft" variants, i.e., quantization, into one dynamic inference framework. Together they achieve 
optimal tradeoffs between accuracy and computational usage by skipping layers if possible or executing varied ``fractions" of the layers. Meanwhile, state-of-the-art hardware design of CNNs have shown that such DFS schemes are  hardware friendly. For example, \citep{Sharma_2018} proposed a bit-flexible CNN accelerator that constitutes an array of bit-level processing elements to dynamically match the bitwidth of each individual layer. With such dedicated accelerators, the proposed DFS's energy savings would be maximized.

\begin{figure}
  \centering
  \includegraphics[width=\columnwidth]{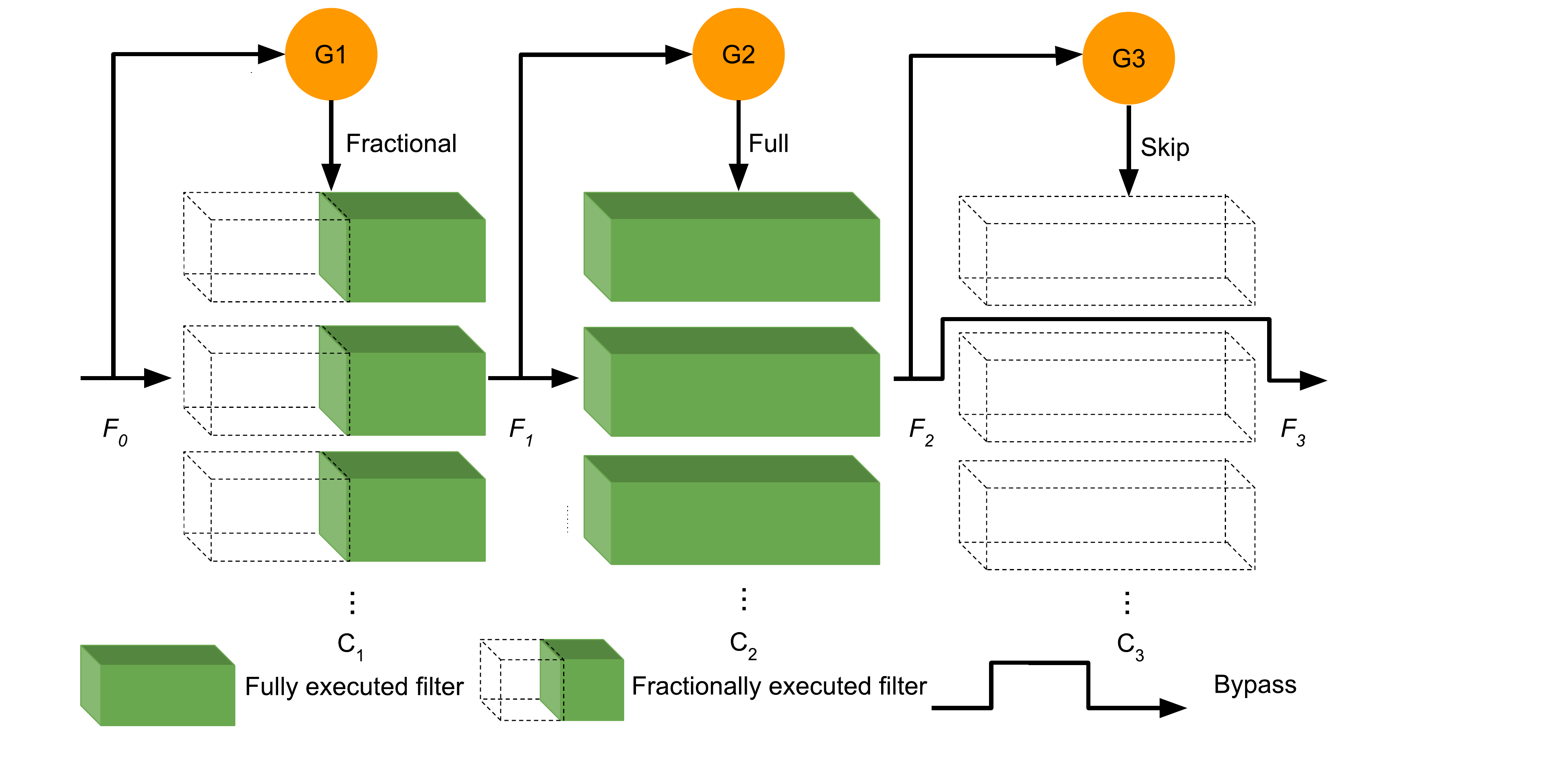}
  \vspace{-2em}
  \caption{An illustration of the DFS framework where C1, C2, C3 denote three consecutive convolution layers, each of which consists of a column of filters as represented using cuboids. In this example, the first conv layer is executed fractionally with a low bitwidth, the second layer is fully executed using the full bitwidth, while the third one is skipped.}

  \label{fig:1}
\end{figure}

\textbf{DFS Overview.} We here introduce how the DFS framework is implemented in ResNet-style models, which has been the most popular backbone CNNs for dynamic inference \citep{wang2018skipnet,wu2018blockdrop}. Figure \ref{fig:1} illustrates the operation of our DFS framework. Specifically, for the $i$-th layer, we let $F_{i} \in R^{s \times s \times m}$ denote its output feature maps and therefore $F_{i-1}$ as its input ones, where $m$ denotes the total number of channels and $s \times s$ denote the feature map size. Also, we employ $C_{i}^k$ to denote the convolutional operation in the $i$-th layer executed in $k$ bits (e.g., $k$ = 32 corresponds to the full bitwidth) and design a gating network $G^{i}$ for determining fractional skipping of the $i$-th layer. Suppose there are a total of $n$ decision options, including the two binary skipping options (i.e., SkipNet) and a set of varied bitwidth options for quantization, and then $G_{i}$ outputs a gating probability vector of length $n$. The operation of a convolutional layer under the DFS framework can then be formulated as:

\vspace{-0.4em}
\begin{equation}
F_{i} = \sum_{k=1}^{n-1} G_{i}^kC_{i}^{b_k}(F_{i-1}) + G_{i}^0F_{i-1}
\label{skipping layer}
\vspace{-0.5em}
\end{equation}

Where $G_{i}$ is the gating probability vector of the ith layer, $G_{i}^k$ denotes the value of its $k$-th entry, and $b_k$ represents the bitwidth option corresponding to the $k$-th entry. When k = 0, we let $G_{i}^0$ represent the probability of a complete skip.  

\begin{figure}
    \centering
    \vspace{-1.5em}
    \includegraphics[width=\columnwidth]{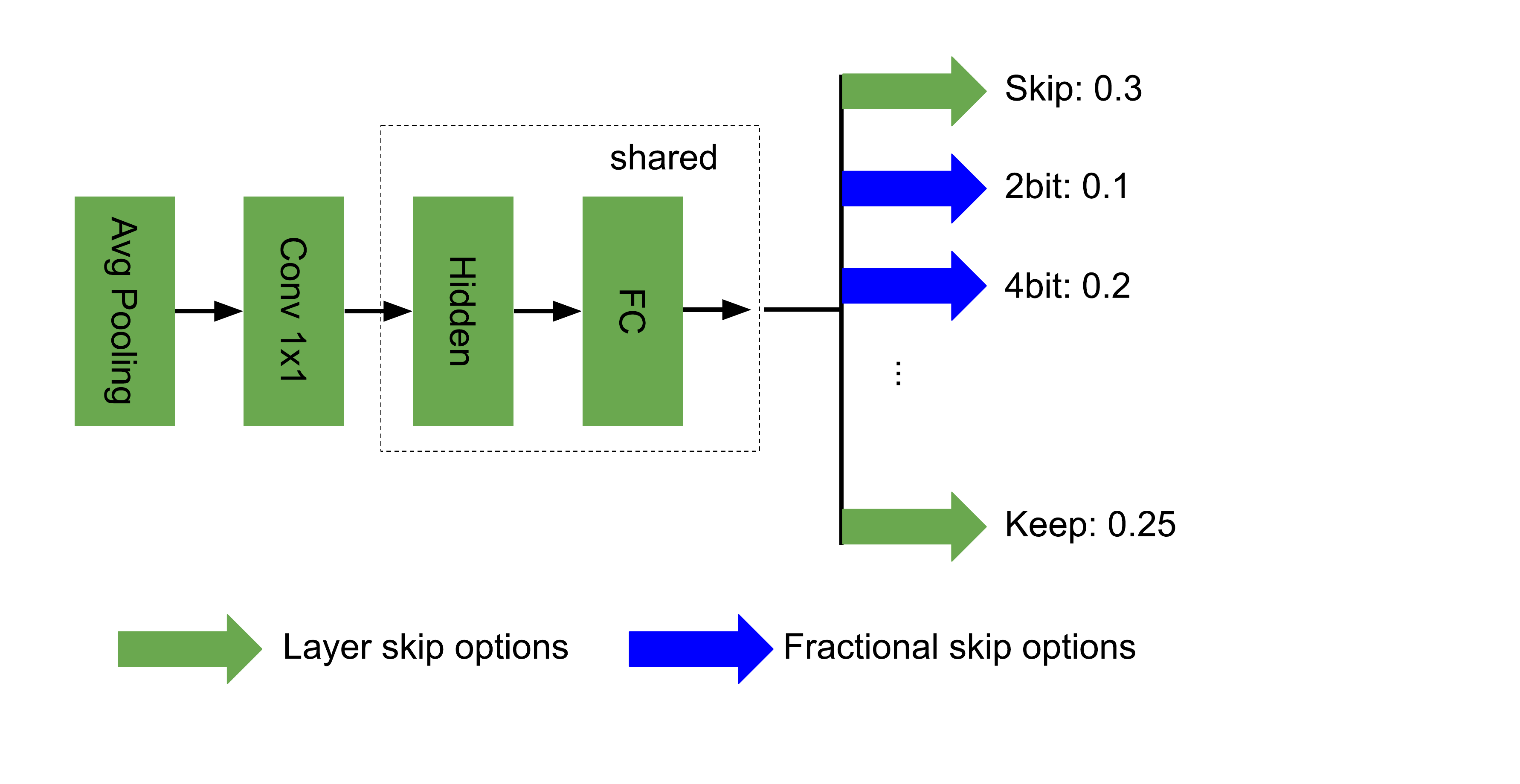}
    \vspace{-3em}
    \caption{An illustration of the RNN gate used in DFS. The output is a skipping probability vector, where the green arrows denote the layer skip options (skip/keep), and the blue arrows represent the quantization options. During inference, the skip/keep/quantization options corresponding to the largest vector element will be selected to be executed.}
    
    \label{fig:2}
\end{figure}

\textbf{Gating Design of DFS.} In the DFS framework, the execution decision of a layer is calculated based on the output of the previous layer. Therefore, the gating network should be able to capture the relevance between consecutive layers in order to make informative decision. As discussed in \citep{wang2018skipnet}, recurrent neural networks (RNNs) have the advantages of both light weight (due to its parameter sharing design, which accounts for only 0.04\% of the computational cost of a residual block) and being able to learn sequential tasks (due to its recurrent structure), thus, we adopt this convention and implement the gating function $G$ as a Long Short Term Memory (LSTM) network, as depicted in Figure \ref{fig:2}. Specifically, suppose there are $n$ options including the binary skipping options and the intermediate bitwidth options, then the LSTM output will be projected into a skipping probability vector of length $n$ via softmax function. During inference, the largest element of the vector will be quantized to 1 and selected for execution; during training, the skipping probability will be used for backpropagation, which will be introduced in more detail in the subsection of DFS training.

\textbf{Training of DFS.}
\underline{Objective Function:} The learning objective of DFS is to boost the prediction accuracy while minimizing the computational cost, and is defined as follows,
\vspace{-0.4em}
\begin{equation}
    \min \limits_{W, G}\,\, L(W, G)\,\, +\, \alpha E(W, G)
    \label{objective function}
\vspace{-0.2em}
\end{equation}

where $L$ represents the accuracy loss, $E$ is the resource-aware loss, $\alpha$ is a weighted factor that trades off importance between accuracy and resource budget, and $W$ and $G$ denote the parameters of the backbone model and the gating network, respectively. The resource-aware loss $E$ is calculated as the total computational cost associated with executed layers (measured in FLOPs in this paper). 

\underline{Skipping Policy Learning with Softmax Approximation: }\\During inference, the execution decision is automatically made based on the skipping probability vector of each gate: \textit{the layer is either skipped or executed in one of the chosen bitwidths}. This \textit{discrete} and therefore \textit{non-differentiable} decision process can make it difficult to train DFS with stochastic gradient descent methods. One alternative is   to use softmax approximation for backpropagation, and quantize to discrete decisions for inference \citep{greff2016highway,wang2018skipnet}. In this paper, we adopt the same technique for training the gating network.    

\underline{Two-Step Training Procedure: }Given a pre-trained CNN model $A$, our goal is to jointly train $A$ and its gating network for targeted computational budget. However, directly training $A$ with randomly initialized gating networks results in much lower accuracy than that of $A$. One possible reason is that with random gate initialization, the model may start at a point with a majority of its layers skipped or executed in low bitwidths, causing large deviation of the feature maps' statistics, which cannot be captured by the batch normalization parameters from the originally trained $A$. To tackle this problem, we conjecture that if the training starts with fully executed model, and then gradually reduces the computational cost towards the target budget, the batch normalization parameters will adapt to the new feature statistics. Motivated by this idea, we use a two-step training procedure:

1) Fix the parameters of $A$ and train the gating network to reach the state of executing all the layers with full bitwidth.

2) With the initialization obtained from the first step, we jointly train $A$ and the gating network to achieve the targeted computational budget.

The computational cost is controlled via \textit{computation percentage} ($cp$), which is defined as the ratio between the FLOPs of executed layers and the FLOPs of the full bitwidth model. During training, we dynamically change the sign of $\alpha$ in in Equation (\ref{objective function}) to stablize the $cp$ of the model: for each iteration, if the $cp$ of the current batch of samples is above the targeted $cp$, we set $\alpha$ to be positive, enforcing the model to reduce its $cp$ by suppressing the resource loss $E$ in Equation (\ref{objective function}); if the $cp$ is below the targeted $cp$, we set $\alpha$ to be negative, encouraging the model to increase its $cp$ by reinforcing $E$. In the end the $cp$ of the model will be stabilized around the targeted $cp$. The absolute value of $\alpha$ is the step size  to adjust the $cp$, since we empirically found out that the performance of our model is robust to a wide range of step sizes, we fix the absolute value of $\alpha$. More detailed experiments of the choice of $\alpha$ will be presented in section \ref{exp}.

\section{Experimental Results}\label{exp}

\textbf{Experiment Setup.} \underline{Models and Datasets:} We evaluate the DFS method using ResNet38 and ResNet74 as the backbone models on two datasets: CIFAR-10 and CIFAR-100. In particular, the structure of the ResNet models follow the design in \citep{he2016deep}. For layer quantization, we consider 4 dynamic execution choices: skip, 8 bits, 16 bits, keep (full bitwidth), and follow the suggestion in \citep{wang2018skipnet} to keep the first residual block always executed with full bitwidth. \underline{Metrics:} We compare DFS with relevant state-of-the-art techniques in terms of the tradeoff between prediction accuracy and computation percentage. Note that for a layer that is executed with a bitwidth of 8 bits, its corresponding computation percentage is $8/32 \times 8/32 = 1/16$ of the full bitwidth layer.

\underline{Training Details:} The training of DFS follows the two-step procedure as described in Section \ref{method}.
For the first step, we set the initial learning rate as 0.1, and train the gating network with a total of 64000 iterations; the learning rate is reduced by $10 \times$ after the 32000-th iteration, and further reduced by $10 \times$ after the 48000-th iteration. The specified computation budget is set to 100\%. The hyperparameters including the momentum, weight decaying factor, and batch size are set to be 0.9, 1e-4, and 128, respectively, and the absolute value of $\alpha$ in Equation (\ref{objective function}) is set to 5e-6.

After the first step is finished, we use the resulting LSTM gating network as the initialization for the second step, where we jointly train the backbone model and gating network to reach the specified computation budget. Here we use an initial learning rate of 0.01, with pre-specified target $cp$, and all other settings are the same as the first step.

\textbf{DFS Performance Evaluation.}\label{performance}
We evaluate the proposed DFS against the competitive dynamic inference technique SkipNet \citep{wang2018skipnet} and two state-of-the-art static CNN quantization techniques proposed in \citep{banner2018scalable} and \citep{wang2019haq}.

\underline{Comparison with SkipNet (Dynamic Inference Method):} In this subsection, we compare the performance of DFS with that of the SkipNet method. Specifically, we compare the performance of DFS on ResNet38 and ResNet74 with that of SkipNet38 and SkipNet74, respectively, on both the CIFAR-10 and CIFAR-100 datasets. We denote DFS-ResNetxx as the models with DFS applied on top of ResNexx backbone.

\begin{figure}[t!]
 
  \centering
  \hspace*{-0.3cm}
  \includegraphics[width=\columnwidth]{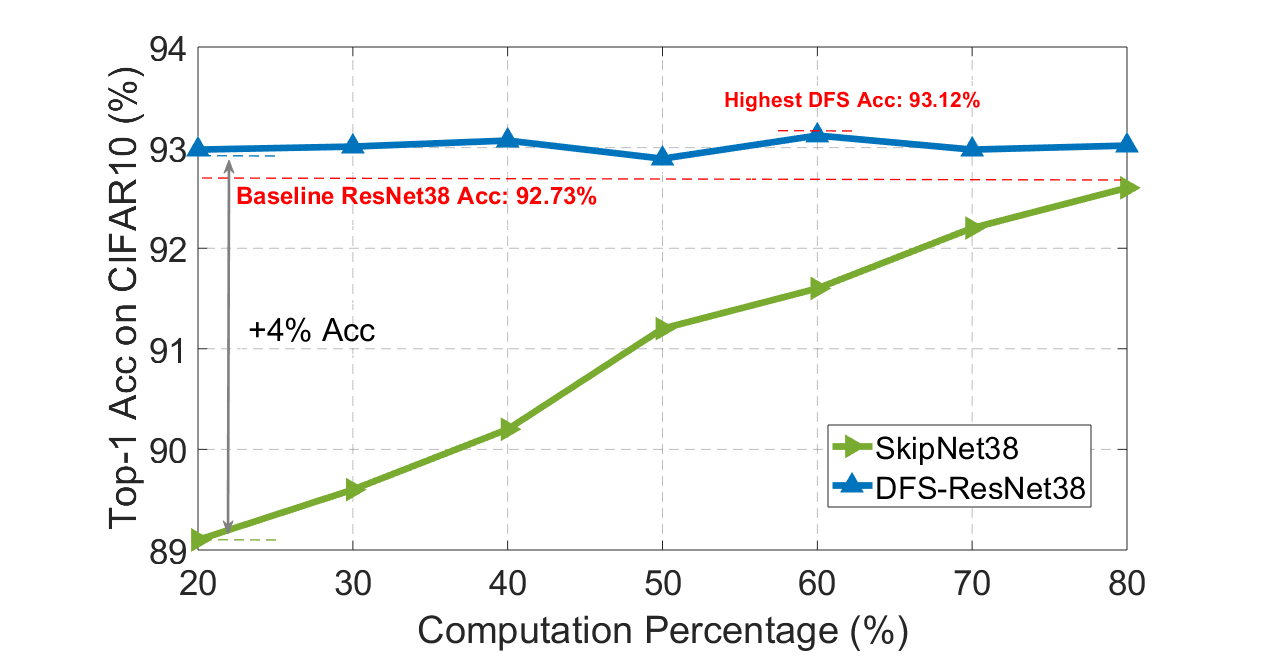}
  \vspace{-1em}
  \caption{Comparing the accuracy vs. computation percentage of DFS-ResNet38 and SkipNet38 on CIFAR10.}
  \vspace{-1em}
  \label{fig:3}
\end{figure}

\begin{figure}[t!]
   \centering
   \hspace*{-0.8cm}
   \includegraphics[width=\columnwidth]{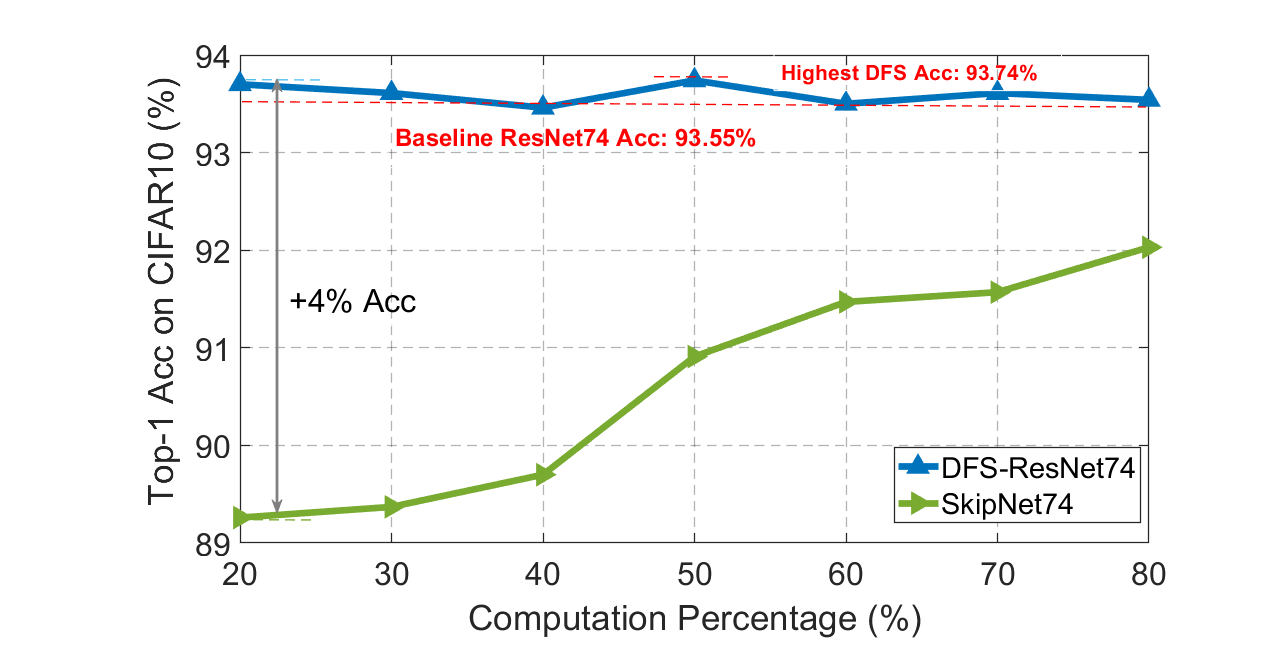}
  \vspace{-1em}
   \caption{Comparing the accuracy vs. computation percentage of DFS-ResNet74 and SkipNet74 on CIFAR10.}
  \vspace{-1em}
   \label{fig:4}
 \end{figure}

Experimental results on \textbf{CIFAR-10} are shown in \textbf{Figure \ref{fig:3}} (vs. ResNet38) and \textbf{Figure \ref{fig:4}} (vs. ResNet74). Specifically, Figures 3-4 show that \textbf{(1) given the same computation budget}, DFS-ResNet38/74 consistently achieves a higher prediction accuracy than that of SkipNet38/74 under a wide range of computation percentage (20\%-80\%), with the largest margin being about ~4\% (93.61\%
vs. 89.26\%) at the computation percentage of 20\%; \textbf{(2) given the same or even with a higher accuracy}, DFS-ResNet38/74 achieves more than 60\% computational saving as compared to SkipNet38/74; \textbf{(3)} interestingly, DFS-ResNet38/74 even achieves better accuracies than the original full bitwidth ResNet38/74. We conjecture that this is because DFS can help in relieving model overfitting thanks to its finer-grained dynamic feature. 

\begin{figure}[t!]
  \centering
  \includegraphics[width=\columnwidth]{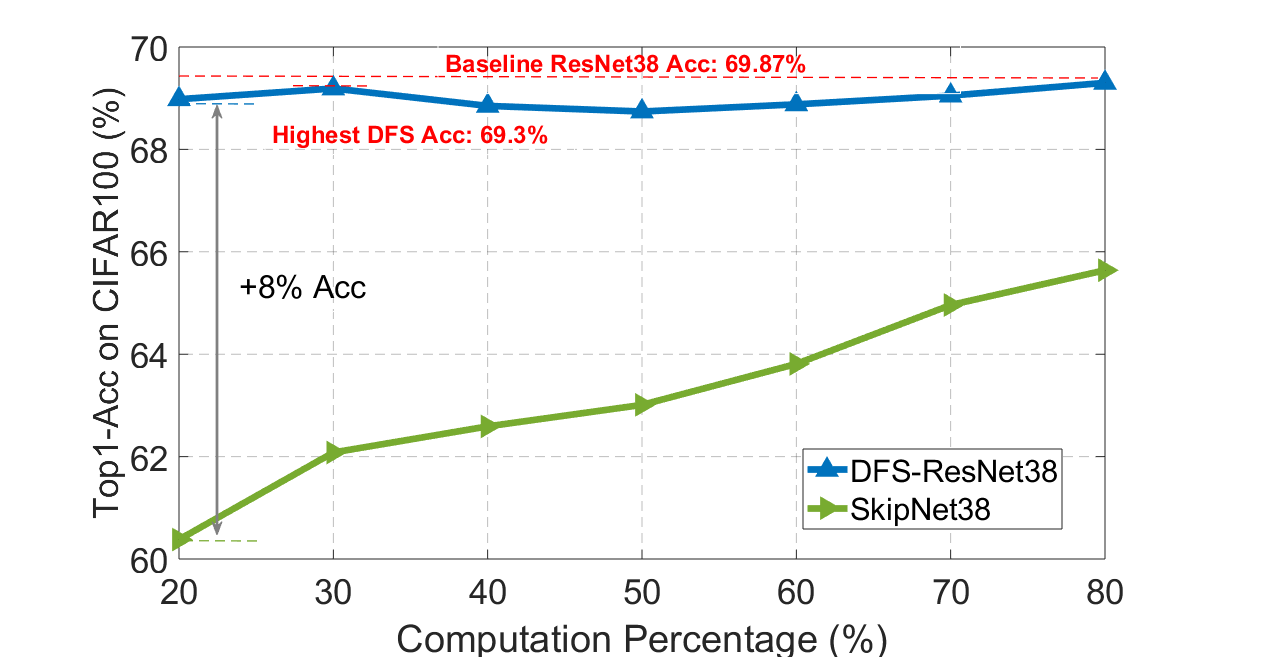}
  \vspace{-2em}
  \caption{Comparing the accuracy vs. computation percentage of DFS-ResNet38 and SkipNet38 on CIFAR-100. }
  \vspace{-1em}
  \label{fig:5}
\end{figure}

\textbf{Figure \ref{fig:5}} (vs. ResNet38) and \textbf{Figure \ref{fig:6}} (vs. ResNet74) show the results on \textbf{CIFAR-100}. We can see that the accuracy improvement (or computational savings) achieved by DFS-ResNet38/74 over SkipNet38/74 is even more pronounced given the same computation percentage (or the same/higher accuracy). For example, as shown in Figure \ref{fig:5}, DFS-ResNet38 achieves 8\% (68.91\% and 60.38\%) better prediction accuracy than SkipNet38 under the computation percentage of 20\%; and Figure \ref{fig:6} shows that DFS-ResNet74 outperforms SkipNet74 with 6\% (70.94\% and 65.09\%)  accuracy when computation percentage is 20\%.

The four sets of experimental results above (i.e., Figures \ref{fig:3}-\ref{fig:6}) show that \textbf{(1)} CNNs with DFS outperform the corresponding SkipNets even at a high computation percentage of 80\% (i.e., small computational savings of 20\% over the original ResNet backbones); \textbf{(2)} as the computation percentage decreases from 80\% to 20\% (corresponding to computational savings from 20\% to 80\%), the prediction accuracy of CNNs with DFS stays relatively stable (slightly fluctuate within a range of 0.5\% while being consistently higher than that of SkipNet under the same computation percentage), whereas the accuracy of SkipNet decreases drastically. These observations validate our conjecture that DFS's finer-grained dynamic execution options can better calibrate the inference accuracy of CNNs w.r.t. the complexity consumed.

\afterpage{
\begin{figure}[!t]
  \centering
  
  \includegraphics[width=\columnwidth]{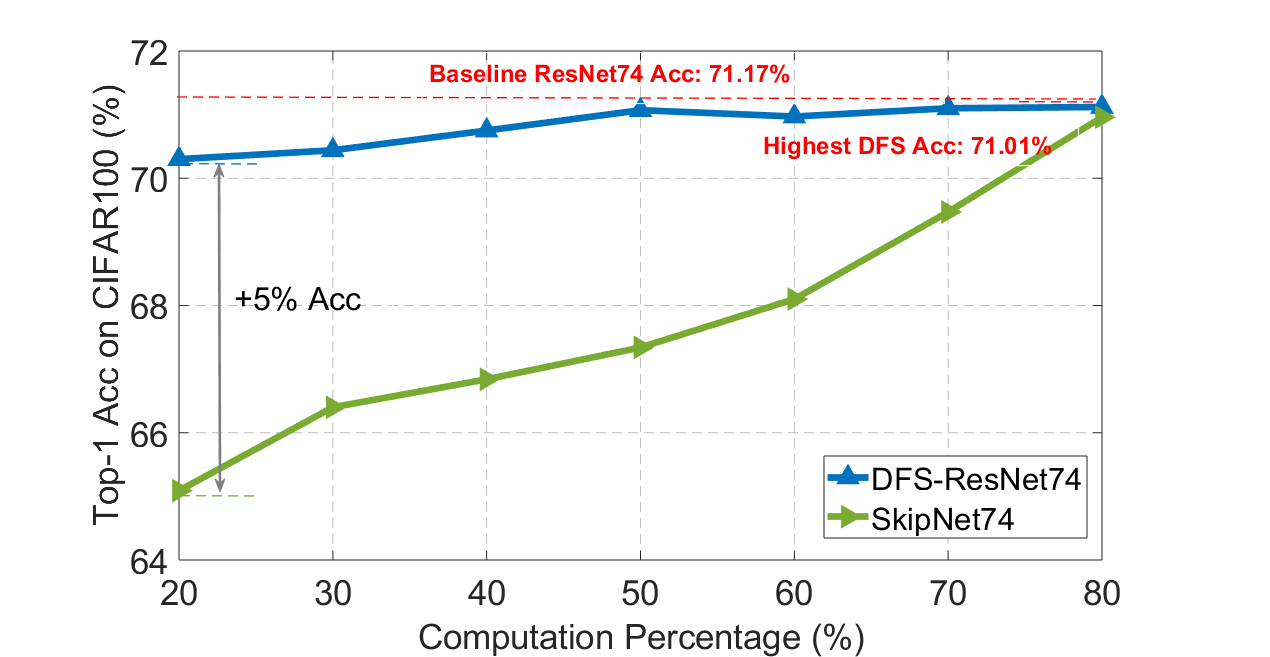}
  \vspace{-2em}
  \caption{Comparing the accuracy vs. computation percentage of DFS-ResNet74 and SkipNet74 on CIFAR-100.}
  \label{fig:6}
\end{figure}
}

\underline{Comparison with Statically Quantized CNNs: }In this section, we compare DFS with two state-of-the-art static CNN quantization methods: the scalable network \citep{banner2018scalable} and HAQ \citep{wang2019haq}, with ResNet38 as the backbone on CIFAR-10. Specifically, for the scalable network \citep{banner2018scalable}, we train it under a set of bitwidths of (8bit, 10bit, 12bit, 14bit, 16bit, 18bit, 20bit, 22bit); according to HAQ's official implementation\footnote{\url{https://github.com/mit-han-lab/haq-release}}, only the weights are quantized, we control HAQ quantized models' computation percentage via compression ratio (the ratio between the size of the quantized weights and the full bitwidth weights). The bitwidth allocation of HAQ is shown in the supplementary material, it can be seen that HAQ learns fine-grained quantization options, the smallest difference between two options is 1 bit.
Note that (1) DFS is orthogonal to static CNN quantization methods, and thus can be applied on top of quantized models for further reducing CNNs' required computational cost; and (2) This set of experiments are not to show that DFS is better than static CNN quantization methods. Instead, the motivation is that it can be insightful to observe the behaviors of static and dynamic quantization methods under the same settings. 

\begin{figure}
  \centering
  \includegraphics[width=\columnwidth]{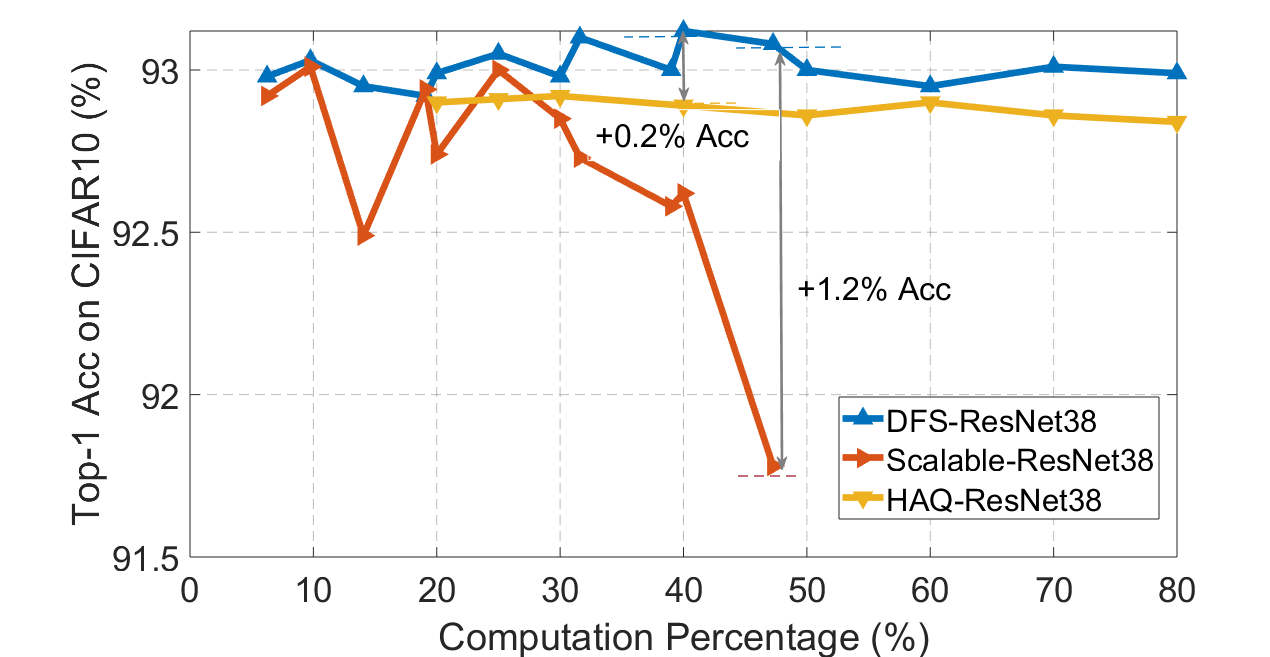}
  \vspace{-2em}
  \caption{Accuracy vs. computation percentage of DFS-ResNet38, the scalable quantized ResNet38 and HAQ quantized ResNet38.}
  \vspace{-1.5em}
  \label{fig:7}
\end{figure}

Figure \ref{fig:7} shows the results. It can be seen that DFS-ResNet38 achieves similar or slightly better accuracy (up to 1.2\% over the scalable method and 0.2\% over HAQ) than both the scalable and HAQ methods, even with a much more coarser-grained quantization options (keep, skip, 8bits, and 16bits). Furthermore, among the three methods, the prediction accuracy of the scalable method fluctuates the most as the computation percentage changes, showing that CNNs with layer-wise adaptive bitwidths can achieve better tradeoffs between accuracy and computational cost.

\underline{Choice of Parameter $\alpha$:} 
We conduct two sets of experiments to demonstrate how dynamically changing the sign of $\alpha$ as described in section \ref{method}
is necessary for reaching the targeted $cp$, and how the absolute value of $\alpha$ affect the model's performance. Table \ref{ablation_alpha} compares two training scenarios of DFS-ResNet74 on CIFAR10, DFS-ResNet74-D denotes the case where dynamic changing sign of $\alpha$ is applied, and DFS-ResNet74-C denotes the case where $\alpha$ is a positive constant, and we set the absolute value of $\alpha$ to 1e-5. It can be seen that when $\alpha$ is constant, the resulting actual $cp$ significantly deviates from the targeted $cp$, since a positive $\alpha$ will keep enforcing the model to redcuce $cp$ without constraint, while the dynamic case achieves the desired $cp$. Table \ref{ablation_abs} shows how the performance of the DFS-ResNet74 model varies with the absolute value of $\alpha$ on CIFAR10. During training, the dynamic changing sing of $\alpha$ is applied. It can be seen that there is an obvious accuracy drop (~0.2\%) under both targeted $cps$ when the absolute value of $\alpha$ increases to (1e-4,1e-3), where the actual $cp$ deviated from the target $cp$ by around 3\%. This is because a larger step size will cause the $cp$ of the model to fluctuate, and thus the unstable training will results in degraded accuracy, while the stable performance in the range (1e-6,1e-5) proves that the model is robust to smaller step size under different targeted $cps$.

\begin{table}[htp!]
\centering
\small
\vspace{-0.5em}
\begin{tabular}{lllll}
    \toprule
    Model & Target $cp $& Actual $cp$ & Acc\\
    \midrule
    DFS-ResNet74-D  & 40\% & 40.20\% & 93.53\% \\
    DFS-ResNet74-C   & 40\% & 8.20\% & 93.12\% \\
    Base-ResNet74    & \- & \- & 93.55\% \\
    \bottomrule
  \end{tabular}
  
\vspace{0.5em}  
\caption{DFS performance under dynamic changing $\alpha$ and constant $\alpha$.}
\vspace{-1em}
\label{ablation_alpha}
\end{table}

\begin{table}[t]
\small
    \centering
    \begin{tabular}{l|cc|cc}
    \toprule
        & \multicolumn{2}{c|}{Target $cp$ = 40 \%}  & \multicolumn{2}{c}{Target $cp$ = 50 \%} \\
        
       abs of $\alpha$ & Actual $cp$  & Acc & Actual $cp$ & Acc \\
    \midrule
       1e-6 & 40.10\% & 93.53\% & 50.08\% & 93.72\%\\
       1e-5 & 40.93\%  & 93.54\% & 50.20\% & 93.74\%\\ 
       1e-4 & 40.80\% & 93.31\% &  51.01\% & 93.52\%\\
       1e-3 & 43.75\% & 93.27\% & 53.40\% & 93.42\%\\

    \bottomrule
    \end{tabular}
    \caption{Performance of DFS models under different $\alpha$. The 'abs' in the leftmost column represents absolute value.}
    \vspace{-1.5em}
    \label{ablation_abs}
\end{table}

\subsection{Decision Behavior Analysis and Visualization}\label{visualization}

We then visualize and study the learned layer-wise decision behaviors of DFS, and how they evolve as $cp$ increases. We demonstrate that quantization options are indeed natural candidates as intermediate ``fractional" skipping choices. Specifically, we investigate how these decisions gradually change to layer quantization at different bitwidths. In general, the (full) layer skip options are likely to be taken only when a very low $cp$ is enforced. When the computational saving requirement is mild, the model shows a strong tendency to  ``fractioally'' skip all its layers.

\begin{figure*}[t!]
  \hspace*{-1.5cm}
  \centering
  \vspace{-2em}
  \includegraphics[width=\textwidth]{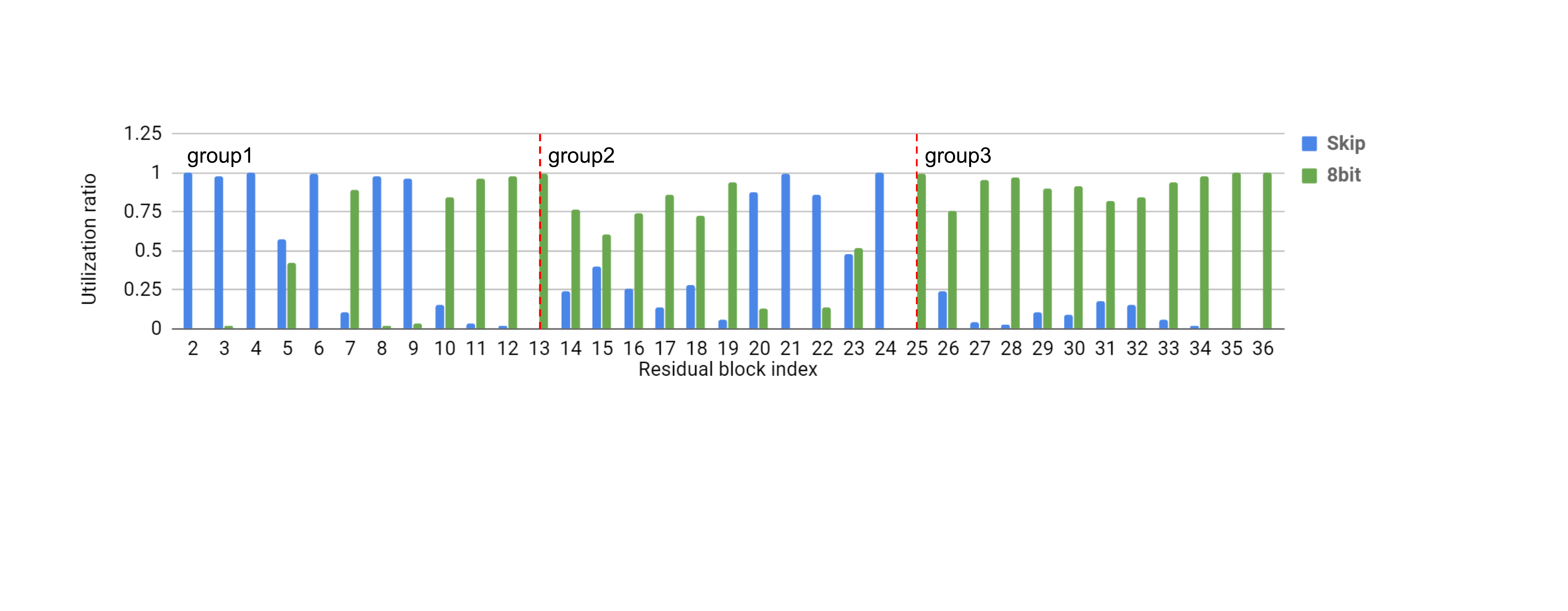}
  \vspace{-6em}
  \caption{Visualization of layerwise decision distribution of DFS-ResNet74 on CIFAR10: computation percentage = 4 \%.}
  \vspace{-1em}
  \label{fig:8}
\end{figure*}

\begin{figure*}[t!]
  \vspace{-8em}
  \hspace*{-1cm}
  \centering
  \includegraphics[width=\textwidth]{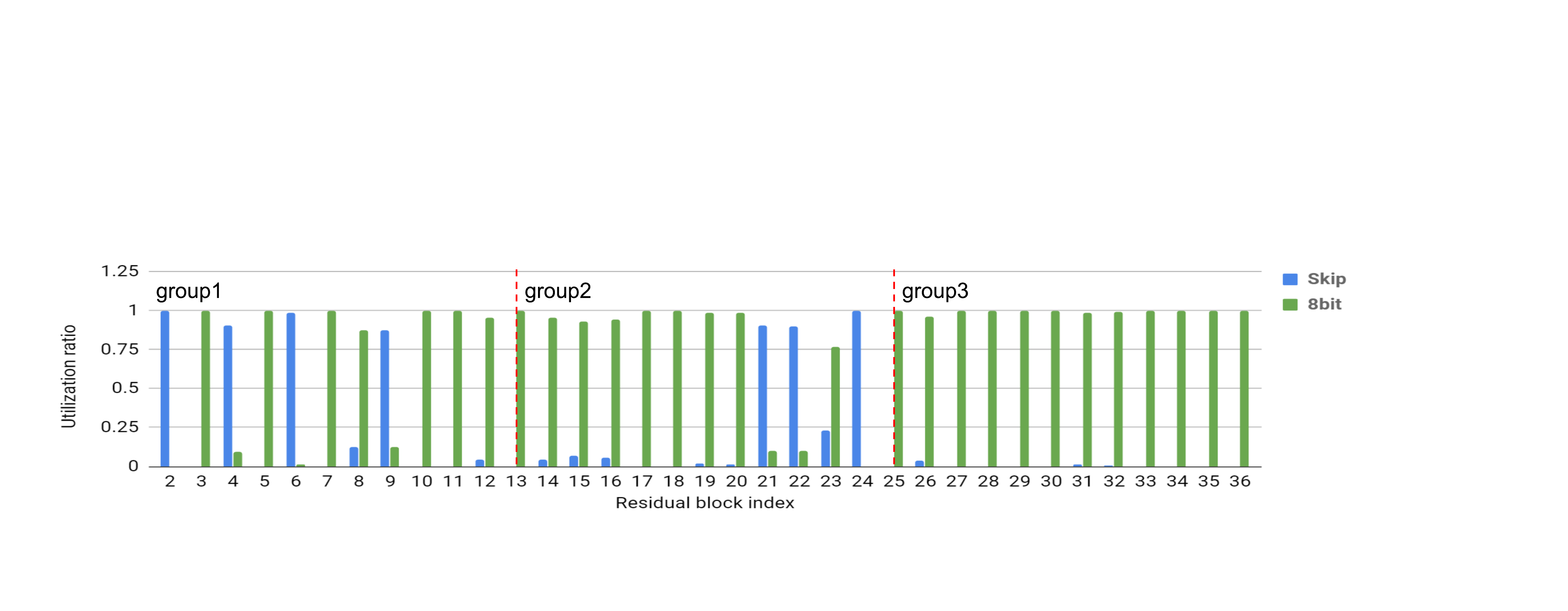}
  \vspace{-4em}
  \caption{Visualization of layerwise decision distribution of DFS-ResNet74 on CIFAR10: computation percentage = 5 \%.}
  \vspace{-1em}
  \label{fig:9}
\end{figure*}

\begin{figure*}[t!]
\vspace{-9em}
  \centering
  \includegraphics[width=\textwidth]{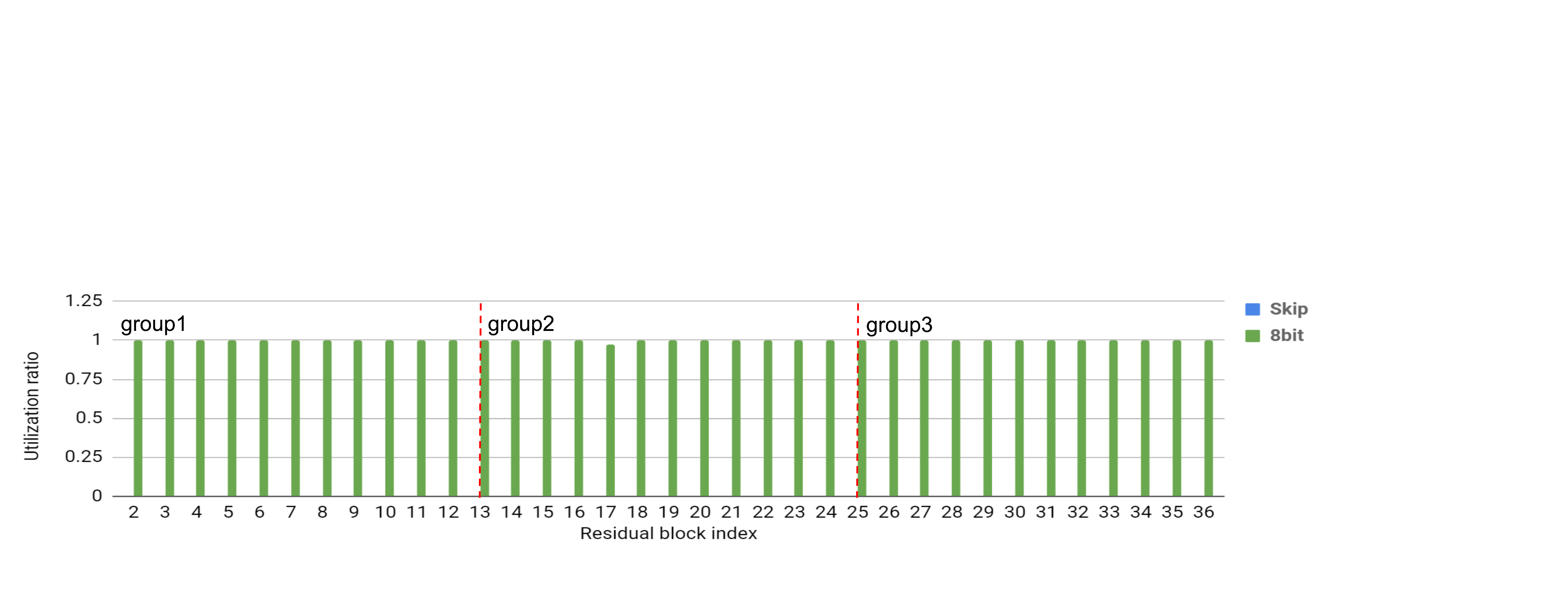}
  \vspace{-4em}
  \caption{Visualization of layerwise decision distribution of DFS-ResNet74 on CIFAR10: computation percentage = 6.25 \%.}
  \vspace{-1em}
  \label{fig:10}
\end{figure*}

Figures \ref{fig:8}-\ref{fig:10} show the layer-wise ``decision distributions'' (e.g., the skip option taken per layer) of DFS-ResNet74 trained on CIFAR-10, as the computation percentage increases from 4\% to 6.25\%. In this specific case and (quite low) percentage range, the model is observed to only choose between ``skip'' and ``8bit'' in a vast majority of input cases. Therefore, we only plot ``skip" and ``8bit" columns for compact display.  we can observe a smooth transition of decision behaviors as the computational percentage varies: from a mixture of layer skipping and quantization, gradually to all layer quantization.  Specifically, from Figure \ref{fig:8} to Figure \ref{fig:9}, within the first residual group, the percentage of skipping options for blocks 2,4,9 remains roughly unchanged, while we observe an obvious drop of skipping percentages at block 5 (from ~55\% to 0\%) and block 8 (from ~100\% to ~10\%). Similarly, for the second and third residual groups, the skipping percentage of most residual blocks gradually reduces to ~0\%, with that of the remaining blocks (20,22,23,24) stays roughly unchanged. From Figure\ref{fig:9} to Figure\ref{fig:10}, the decisions of all the layers shift to 8bit. The smooth transition empirically endorses our hypothesis made in Section \ref{method}, that the layer quantization options can serve as a ``fractional" intermediate stage between the binary layer skipping options.

\begin{figure}[t!]
  \centering
  
  \includegraphics[width=\columnwidth]{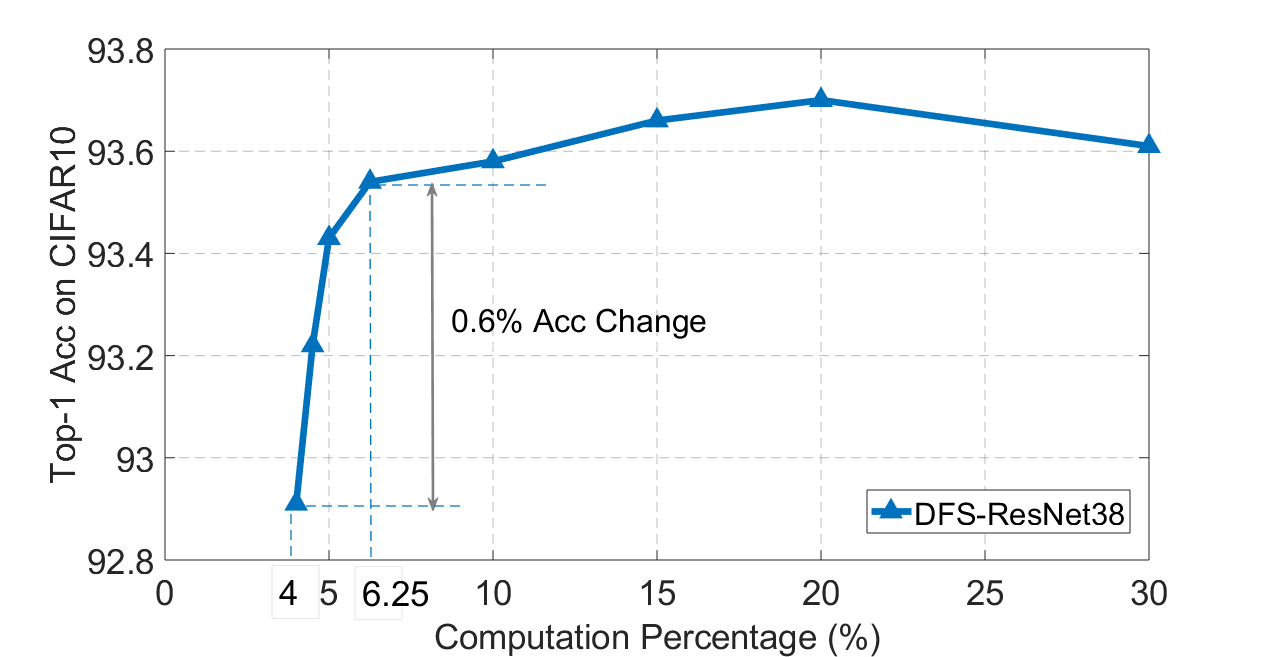}
  \vspace{-2em}
  \caption{DFS-ResNet74 under lower $cps$.}
  \vspace{-1em}
  \label{fig:11}
\end{figure}

As $cp$ increases, DFS apparently favors the finer-grained layer quantization options than the coarser-grained layer skipping. Figure \ref{fig:11} shows the accuracy of DFS-ResNet74 when the computation percentage increases from 4\% to 30\%. From 4\% to 6.25\% (when the layer skipping options gradually change to all-quantization options), there is a notable accuracy increase from 92.91\% to 93.54\%. The performance then reaches a plateau after the computation percentage of 6.25\%, while we observe DFS now tends to choose quantization for all layers (see supplementary material). This phenomenon demonstrates that as a ``fractional" layer skipping candidate, low bitwidth options can better restore the model's accuracy under a wider range of $cps$.

Additionally, from Figures \ref{fig:8}-\ref{fig:10}, we observe that the first residual block within each residual group is learned not to be skipped, regardless of the values of $cps$. That aligns with the findings in \citep{greff2016highway}, which shows that for ResNet-style models, only the first residual block within each group extracts a completely new representation (and therefore being most important), and that the remaining residual blocks within the same group only refine this feature iteratively. 

\section{Conclusion}
We proposed a novel DFS framework, which extends binary layer skipping options with the ``fractional skipping'' ability - by quantizing the layer weights and activations into different bitwidths. The DFS framework exploits model redundancy in a much finer-grained level, leading to more flexible and effective calibration between inference accuracy and complexity. We evaluate DFS on the CIFAR-10 and CIFAR-100 benchmarks, and it was shown to compare favorably against both state-of-the-art dynamic inference method and static quantization techniques. 

While we demonstrate that quantization indeed can be viewed as ``fractional" intermediate states in-between binary layer skip options (by both achieved results,  and the visualizations of skipping decision transitions), we recognize that more possible alternatives to ``fractionally'' execute a layer could be explored, such as channel slimming \citep{yu2018slimmable}. We leave this as a future work.

\bibliographystyle{aaai}
\bibliography{aaai20}

\begin{thebibliography}{}

\bibitem[\protect\citeauthoryear{Banner \bgroup et al\mbox.\egroup
  }{2018}]{banner2018scalable}
Banner, R.; Hubara, I.; Hoffer, E.; and Soudry, D.
\newblock 2018.
\newblock Scalable methods for 8-bit training of neural networks.
\newblock In {\em NIPS}.

\bibitem[\protect\citeauthoryear{Chen \bgroup et al\mbox.\egroup
  }{2018}]{chen2018gaternet}
Chen, Z.; Li, Y.; Bengio, S.; and Si, S.
\newblock 2018.
\newblock Gaternet: Dynamic filter selection in convolutional neural network
  via a dedicated global gating network.
\newblock {\em arXiv preprint arXiv:1811.11205}.

\bibitem[\protect\citeauthoryear{Cheng \bgroup et al\mbox.\egroup
  }{2017}]{cheng2017survey}
Cheng, Y.; Wang, D.; Zhou, P.; and Zhang, T.
\newblock 2017.
\newblock A survey of model compression and acceleration for deep neural
  networks.
\newblock {\em arXiv preprint arXiv:1710.09282}.

\bibitem[\protect\citeauthoryear{Figurnov \bgroup et al\mbox.\egroup
  }{2017}]{figurnov2017spatially}
Figurnov, M.; Collins, M.~D.; Zhu, Y.; Zhang, L.; Huang, J.; Vetrov, D.; and
  Salakhutdinov, R.
\newblock 2017.
\newblock Spatially adaptive computation time for residual networks.
\newblock In {\em CVPR}.

\bibitem[\protect\citeauthoryear{Greff, Srivastava, and
  Schmidhuber}{2016}]{greff2016highway}
Greff, K.; Srivastava, R.~K.; and Schmidhuber, J.
\newblock 2016.
\newblock Highway and residual networks learn unrolled iterative estimation.
\newblock {\em arXiv preprint arXiv:1612.07771}.

\bibitem[\protect\citeauthoryear{Han \bgroup et al\mbox.\egroup
  }{2015}]{han2015learning}
Han, S.; Pool, J.; Tran, J.; and Dally, W.
\newblock 2015.
\newblock Learning both weights and connections for efficient neural network.
\newblock In {\em NIPS}.

\bibitem[\protect\citeauthoryear{Han, Mao, and Dally}{2015}]{han2015deep}
Han, S.; Mao, H.; and Dally, W.~J.
\newblock 2015.
\newblock Deep compression: Compressing deep neural networks with pruning,
  trained quantization and huffman coding.
\newblock {\em arXiv preprint arXiv:1510.00149}.

\bibitem[\protect\citeauthoryear{He \bgroup et al\mbox.\egroup
  }{2016}]{he2016deep}
He, K.; Zhang, X.; Ren, S.; and Sun, J.
\newblock 2016.
\newblock Deep residual learning for image recognition.
\newblock In {\em CVPR}.

\bibitem[\protect\citeauthoryear{He, Zhang, and Sun}{2017}]{he2017channel}
He, Y.; Zhang, X.; and Sun, J.
\newblock 2017.
\newblock Channel pruning for accelerating very deep neural networks.
\newblock In {\em ICCV}.

\bibitem[\protect\citeauthoryear{Huang \bgroup et al\mbox.\egroup
  }{2017}]{huang2017multi}
Huang, G.; Chen, D.; Li, T.; Wu, F.; van~der Maaten, L.; and Weinberger, K.~Q.
\newblock 2017.
\newblock Multi-scale dense networks for resource efficient image
  classification.
\newblock {\em arXiv preprint arXiv:1703.09844}.

\bibitem[\protect\citeauthoryear{Jacob \bgroup et al\mbox.\egroup
  }{2018}]{jacob2018quantization}
Jacob, B.; Kligys, S.; Chen, B.; Zhu, M.; Tang, M.; Howard, A.; Adam, H.; and
  Kalenichenko, D.
\newblock 2018.
\newblock Quantization and training of neural networks for efficient
  integer-arithmetic-only inference.
\newblock In {\em CVPR}.

\bibitem[\protect\citeauthoryear{Kim, Ahn, and Oh}{2018}]{kim2018nestednet}
Kim, E.; Ahn, C.; and Oh, S.
\newblock 2018.
\newblock Nestednet: Learning nested sparse structures in deep neural networks.
\newblock In {\em CVPR}.

\bibitem[\protect\citeauthoryear{Krizhevsky, Sutskever, and
  Hinton}{2012}]{krizhevsky2012imagenet}
Krizhevsky, A.; Sutskever, I.; and Hinton, G.~E.
\newblock 2012.
\newblock Imagenet classification with deep convolutional neural networks.
\newblock In {\em NIPS}.

\bibitem[\protect\citeauthoryear{Lin \bgroup et al\mbox.\egroup
  }{2017}]{lin2017runtime}
Lin, J.; Rao, Y.; Lu, J.; and Zhou, J.
\newblock 2017.
\newblock Runtime neural pruning.
\newblock In {\em NIPS}.

\bibitem[\protect\citeauthoryear{Liu and Deng}{2018}]{liu2018dynamic}
Liu, L., and Deng, J.
\newblock 2018.
\newblock Dynamic deep neural networks: Optimizing accuracy-efficiency
  trade-offs by selective execution.
\newblock In {\em AAAI}.

\bibitem[\protect\citeauthoryear{Liu \bgroup et al\mbox.\egroup
  }{2017}]{liu2017learning}
Liu, Z.; Li, J.; Shen, Z.; Huang, G.; Yan, S.; and Zhang, C.
\newblock 2017.
\newblock Learning efficient convolutional networks through network slimming.
\newblock In {\em ICCV}.

\bibitem[\protect\citeauthoryear{Rastegari \bgroup et al\mbox.\egroup
  }{2016}]{rastegari2016xnor}
Rastegari, M.; Ordonez, V.; Redmon, J.; and Farhadi, A.
\newblock 2016.
\newblock Xnor-net: Imagenet classification using binary convolutional neural
  networks.
\newblock In {\em ECCV}.

\bibitem[\protect\citeauthoryear{Sharma \bgroup et al\mbox.\egroup
  }{2018}]{Sharma_2018}
Sharma, H.; Park, J.; Suda, N.; Lai, L.; Chau, B.; Chandra, V.; and
  Esmaeilzadeh, H.
\newblock 2018.
\newblock Bit fusion: Bit-level dynamically composable architecture for
  accelerating deep neural network.
\newblock {\em 2018 ACM/IEEE 45th Annual (ISCA)}.

\bibitem[\protect\citeauthoryear{Taigman \bgroup et al\mbox.\egroup
  }{2014}]{taigman2014deepface}
Taigman, Y.; Yang, M.; Ranzato, M.; and Wolf, L.
\newblock 2014.
\newblock Deepface: Closing the gap to human-level performance in face
  verification.
\newblock In {\em CVPR}.

\bibitem[\protect\citeauthoryear{Teerapittayanon, McDanel, and
  Kung}{2016}]{teerapittayanon2016branchynet}
Teerapittayanon, S.; McDanel, B.; and Kung, H.-T.
\newblock 2016.
\newblock Branchynet: Fast inference via early exiting from deep neural
  networks.
\newblock In {\em 2016 23rd ICPR}.

\bibitem[\protect\citeauthoryear{Teja~Mullapudi \bgroup et al\mbox.\egroup
  }{2018}]{teja2018hydranets}
Teja~Mullapudi, R.; Mark, W.~R.; Shazeer, N.; and Fatahalian, K.
\newblock 2018.
\newblock Hydranets: Specialized dynamic architectures for efficient inference.
\newblock In {\em CVPR}.

\bibitem[\protect\citeauthoryear{Wang \bgroup et al\mbox.\egroup
  }{2018a}]{wang2018skipnet}
Wang, X.; Yu, F.; Dou, Z.-Y.; Darrell, T.; and Gonzalez, J.~E.
\newblock 2018a.
\newblock Skipnet: Learning dynamic routing in convolutional networks.
\newblock In {\em ECCV}.

\bibitem[\protect\citeauthoryear{Wang \bgroup et al\mbox.\egroup
  }{2018b}]{NIPS-18}
Wang, Y.; Nguyen, T.; Zhao, Y.; Wang, Z.; Lin, Y.; and Baraniuk, R.
\newblock 2018b.
\newblock Energynet: Energy-efficient dynamic inference.
\newblock In {\em Thirty-second Conference on Neural Information Processing
  Systems (NIPS 2018) Workshop}.

\bibitem[\protect\citeauthoryear{Wang \bgroup et al\mbox.\egroup
  }{2019}]{wang2019haq}
Wang, K.; Liu, Z.; Lin, Y.; Lin, J.; and Han, S.
\newblock 2019.
\newblock Haq: Hardware-aware automated quantization with mixed precision.
\newblock In {\em CVPR}.

\bibitem[\protect\citeauthoryear{Wen \bgroup et al\mbox.\egroup
  }{2016}]{wen2016learning}
Wen, W.; Wu, C.; Wang, Y.; Chen, Y.; and Li, H.
\newblock 2016.
\newblock Learning structured sparsity in deep neural networks.
\newblock In {\em NIPS}.

\bibitem[\protect\citeauthoryear{Wu \bgroup et al\mbox.\egroup
  }{2018a}]{wu2018deep}
Wu, J.; Wang, Y.; Wu, Z.; Wang, Z.; Veeraraghavan, A.; and Lin, Y.
\newblock 2018a.
\newblock Deep k-means: Re-training and parameter sharing with harder cluster
  assignments for compressing deep convolutions.
\newblock In {\em International Conference on Machine Learning},  5359--5368.

\bibitem[\protect\citeauthoryear{Wu \bgroup et al\mbox.\egroup
  }{2018b}]{wu2018blockdrop}
Wu, Z.; Nagarajan, T.; Kumar, A.; Rennie, S.; Davis, L.~S.; Grauman, K.; and
  Feris, R.
\newblock 2018b.
\newblock Blockdrop: Dynamic inference paths in residual networks.
\newblock In {\em CVPR}.

\bibitem[\protect\citeauthoryear{Xu, Park, and Brick}{2018}]{xu2018hybrid}
Xu, X.; Park, M.~S.; and Brick, C.
\newblock 2018.
\newblock Hybrid pruning: Thinner sparse networks for fast inference on edge
  devices.
\newblock {\em arXiv preprint arXiv:1811.00482}.

\bibitem[\protect\citeauthoryear{Yu \bgroup et al\mbox.\egroup
  }{2017}]{yu2017scalpel}
Yu, J.; Lukefahr, A.; Palframan, D.; Dasika, G.; Das, R.; and Mahlke, S.
\newblock 2017.
\newblock Scalpel: Customizing dnn pruning to the underlying hardware
  parallelism.
\newblock {\em ACM SIGARCH Computer Architecture News}.

\bibitem[\protect\citeauthoryear{Yu \bgroup et al\mbox.\egroup
  }{2018}]{yu2018slimmable}
Yu, J.; Yang, L.; Xu, N.; Yang, J.; and Huang, T.
\newblock 2018.
\newblock Slimmable neural networks.
\newblock {\em arXiv preprint arXiv:1812.08928}.

\bibitem[\protect\citeauthoryear{Zhou \bgroup et al\mbox.\egroup
  }{2016}]{zhou2016dorefa}
Zhou, S.; Wu, Y.; Ni, Z.; Zhou, X.; Wen, H.; and Zou, Y.
\newblock 2016.
\newblock Dorefa-net: Training low bitwidth convolutional neural networks with
  low bitwidth gradients.
\newblock {\em arXiv preprint arXiv:1606.06160}.

\bibitem[\protect\citeauthoryear{Zhu \bgroup et al\mbox.\egroup
  }{2016}]{zhu2016trained}
Zhu, C.; Han, S.; Mao, H.; and Dally, W.~J.
\newblock 2016.
\newblock Trained ternary quantization.
\newblock {\em arXiv preprint arXiv:1612.01064}.

\end{thebibliography}

\end{document}